# CLOSE-UP VIEW SYNTHESIS BY INTERPOLATING OPTICAL FLOW

*Xinyi Bai, Ze Wang, Lu Yang, Hong Cheng*

Center for Robotics, University of Electronic Science and Technology of China, Chengdu, China
wwwbxy123@163.com, 254635826@qq.com, {yanglu, hcheng}@uestc.edu.cn

## ABSTRACT

The virtual viewpoint is perceived as a new technique in virtual navigation, as yet not supported due to the lack of depth information and obscure camera parameters. In this paper, a method for achieving close-up virtual view is proposed and it only uses optical flow to build parallax effects to realize pseudo 3D projection without using depth sensor. We develop a bidirectional optical flow method to obtain any virtual viewpoint by proportional interpolation of optical flow. Moreover, with the ingenious application of the optical-flow-value, we achieve clear and visual-fidelity magnified results through lens stretching in any corner, which overcomes the visual distortion and image blur through viewpoint magnification and transition in Google Street View system.

## 1. INTRODUCTION

There has been a lot of interest from the computer graphics and computer vision fields towards image-based modeling and rendering methods (IBR) [1]. Virtual viewpoint is an important application of such research, including virtual tours of interest places, close-up virtual point between real-scenes. Any kind of application where achieving photorealism is a goal can actually benefit from advances in this field [2]. Recent virtual view applications use collections of images in purpose of allowing users to remotely explore a given existing environment and magnify these pictures according to the scene transition or customers' demand. The Google Street View system (GSV) is a good example of a large-scale image-based model of city landscape. The system by Furukawa et al. [3] is another example of virtual navigation system.

 Image Based Rendering (IBR) is a new method that has been widely used in creating virtual viewpoint scenes. It is preferred over the traditional 3D geometry modeling and rendering technology that has limitations on rendering quality especially in magnified areas and scene complexity. Also IBR has a potential to create close-up output images if the magnified rendering stage can preserve the visual fidelity of the reference images.

Optical flow is the distribution of apparent velocities of movement of brightness patterns in an image [4]. Not only it contains rich 2D motion cues, but also contain 3D scene structure information [5]. Recently, optical flow estimation has been the foundation of many technologies for motion estimation, tracking and image segmentation [6], which can acquire motion cues of targets under scene changes[7], and gain target's zone and scale cues if deeper analysis is done. Classical optical flow algorithms [4, 8, 9] use pixel value to build targets, compute optical flow under brightness constancy assumption and smoothness restriction. But in real motion analysis, the brightness constancy assumption that pixel values are invariant from one frame to another along with the motion, often breaks because of illumination changes and noise [10], especially for the relatively static part of moving objects with small chromatic aberration. It is difficult to acquire a precise current motion information in these parts from single optical flow.

And traditional virtual viewpoint used in 3D city modelling like Google Street View system [11] is poor in close-up scenes. Because it will cause visual unreality just using simple magnification on the zone of detail we are interested in, which cannot be compared with 3D projection [13] based on depth of field [14], briefly, it has poor close-up effects. In view of the large cost and huge computational load of depth sensor, in this paper, a novel close-up scene method is proposed aiming to use optical flow to construct parallax, finally, realize pseudo 3D projection and overcomes the defect of traditional IBR technology, such as the visual unreality caused by simple magnification.

## 2. CAMERA CALIBRATION

### 2.1. Geometric registration

Due to the estimation error of internal and external camera parameters and different view angles, the initial two images are not in the same plane, which make the optical flow and virtual viewpoint images difficult to obtain. Thus, geometric registration [12] is required to project them to the same plane.

In computer vision, it is common to calculate homography H [12] by finding the transfer equation from world coordinate to image coordinate as:

$$Z_C \begin{bmatrix} u \\ v \\ 1 \end{bmatrix} = M_1 M_2 A_w = H A_w \quad (1)$$

where $A_w(x_w, y_w, z_w)^T$ is point A in the world coordinate, $(u, v)^T$ is the aligned point in image coordinate, $Z_c$ is the Z axis component of real object in the panoramic imaging plane, $M_1$ is the internal camera parameter matrix, $M_2$ is the external camera parameter matrix. A more detailed form is described as:

$$Z_C \begin{bmatrix} u \\ v \\ 1 \end{bmatrix} = \begin{bmatrix} 1/d_x & 0 & u_0 \\ 0 & 1/d_y & v_0 \\ 0 & 0 & 1 \end{bmatrix} \begin{bmatrix} f & 0 & 0 & 0 \\ 0 & f & 0 & 0 \\ 0 & 0 & 1 & 0 \end{bmatrix} \begin{bmatrix} R & T \\ 0^T & 1 \end{bmatrix} \begin{bmatrix} x_w \\ y_w \\ z_w \\ 1 \end{bmatrix}$$

$$= \begin{bmatrix} \alpha & 0 & u_0 & 0 \\ 0 & \beta & v_0 & 0 \\ 0 & 0 & 1 & 0 \end{bmatrix} \begin{bmatrix} R & T \\ 0^T & 1 \end{bmatrix} \begin{bmatrix} x_w \\ y_w \\ z_w \\ 1 \end{bmatrix} = M_1 M_2 A_w = H A_w \quad (2)$$

We project the left and right images into the world coordinate respectively, deduce $H_1$ and $H_2$. Then calculate the homography matrix $H'$ to correct the two images in the same plane, finally completing the geometric registration.

$$z_{c1} \begin{bmatrix} u_1 \\ v_1 \\ 1 \end{bmatrix} = H_1 A_w \quad z_{c2} \begin{bmatrix} u_2 \\ v_2 \\ 1 \end{bmatrix} = H_2 A_w' \quad \begin{bmatrix} u_2 \\ v_2 \\ 1 \end{bmatrix} = H \begin{bmatrix} u_1 \\ v_1 \\ 1 \end{bmatrix}$$

$$H' = H_1 \cdot H_2^{-1} \quad (3)$$

Where $[u_1 \ v_1 \ 1]^T$ is a point in the left image coordinate, while $[u_2 \ v_2 \ 1]^T$ is the aligned point in the right image coordinate.

### 2.2. Color correction

Another crucial step in optical flow calculation is to do the color correction [15] in purpose of maintaining the color constancy and unity.

The RGB pixel value of point $(x, y)$ in the source image is $I_1(x, y)$, and the corrected pixel value is $I_2(x, y)$, in order to avoid the result to be too bright or dark, this paper uses a 3-sections piecewise function to correct the pixel value, as shown in **Figure 1.**
The pixel value transfer function is defined as:

$$I_2(x, y) = F(I_1(x, y)) \quad (4)$$

In zone $(m1 - m2)$ the transfer function is a linear function. So it can be defined as:

$$F(x) = a + bx \quad (5)$$

In zone $(0 - m1)$ and $(m2 - 255)$ the transfer function is a $\gamma$ function:

$$F(x) = a + k(x/k)^b \quad (6)$$

In Figure 1.where pixel value of m1 is 5% of the total histogram of source image and m2 is 95 % of its total histogram. By using the given criterion of m1 and m2, color correction can be achieved.

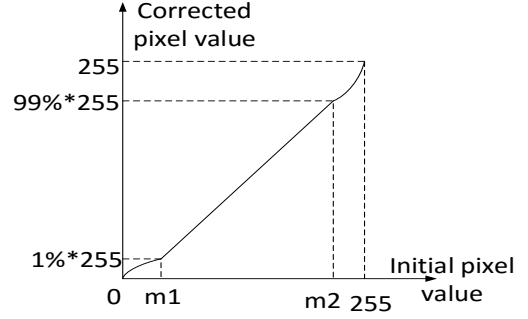

**Figure 1.** The pixel value transfer function

### 3. CLOSE-UP VIRTUAL VIEWPOINT GENERATION

Any close-up viewpoint can be generated directly by optical flow without the aid of other image or any auxiliary tools. In traditional methods, occlusion is a significant problem, our method use bidirectional optical flow and select the value of pixel weight to solve it, at the same time it also deals with the distortion at the edge of image caused by the error of unidirectional optical flow.

First we take two pictures (image 1, image 2) from different view angles, then obtain the bidirectional optical flow $I_1$ and $I_2$ between image 1 and image 2. To get a viewpoint (image 3) which is collinear with image 1 and image 2, we interpolate the optical flow $I_2$ to get optical flow $I_3$ from image 1 to image 4, and interpolate the optical flow $I_3$ to get optical flow $I_4$ from image 2 to image 5, then with the use of optical flow $I_3$ and $I_4$, we can project our source image 1 to virtual image 4 and source image 2 to virtual image 5 respectively. Finally by fusing image 4 and image 5, we can get a high quality image 3, the process is shown in Figure 2.

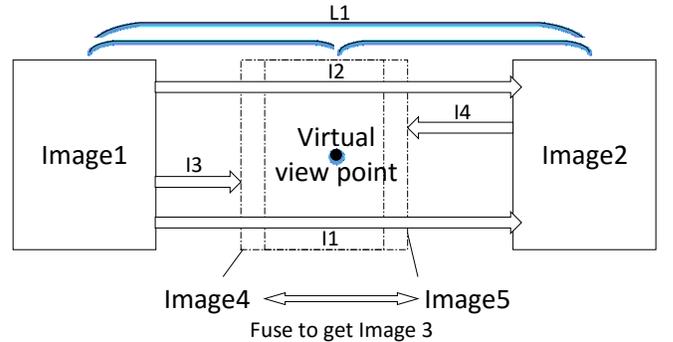

**Figure 2**. The sketch map of virtual viewpoint

Our challenging work is how to calculate the optical flow $I_3$ and $I_4$. Our method is using the fact that the ratio of physical distance must be in proportion to the ratio of optical flow. Thus, by calculating the physical distance $L_1$ between image 1 and image 2, $L_2$ between virtual viewpoint image 4 and image 1, and $L_3$ between virtual viewpoint image 5 and image 2. Set $a = \frac{L_2}{L_1}$, $b = \frac{L_3}{L_1}$, so $I_3 = aI_1$, $I_4 = bI_2$.

Hence, virtual image 4 and image 5 are determined by $I_3$ and $I_4$ respectively, finally by fusing the two virtual images, image 3 from virtual viewpoint can be achieved.

With optical flow interpolation, we realize virtual viewpoint. Then we can use optical flow to get a close-up image. Compared with GSV system' lens stretch, traditional simple magnification without depth will cause visual distortion, because image magnification is limited to 2 dimensional. Considering depth information plays a decisive role in the process of lens stretching, while the depth of field is unreachable for most ordinary cameras. Our main contribution is to present a pseudo 3D projection magnification method based on optical flow.

Our method is based on the theory that the value of optical flow has inverse relation with depth at each pixel, the greater optical flow value implies the smaller the depth of field. That inspires us to use optical flow to construct parallax which indirectly replaces the depth of field. In the same way, given that the ratio of depth, equals to the ratio of optical-flow-value, it can be used to separate the target from the background.

Considering the error of optical flow, we applied circle graph method for the area around the optical flow field. And by selecting the optical flow in the relevant region, we get the integrity foreground with the same depth, which can deal with the defect of the miscalculated parallax caused by the error of optical flow. For example, in a moving object, the value of optical flow in the relatively static part may be 0 in experiments, it is particularly obvious for image with small chromatic aberration. To avoid these miscalculation, we choose the close area which is enclosed by the edge of the moving target as foreground.

With our parallax calculated from optical flow, combined with $I_3$, $I_4$ from section 3, we can get two close-up images, finally by fusing the two close-up images, our close-up result is achieved. Our method use optical flow to get parallax, which can largely reduce the stitching cost. The sketch map of this process is shown in Figure 3.

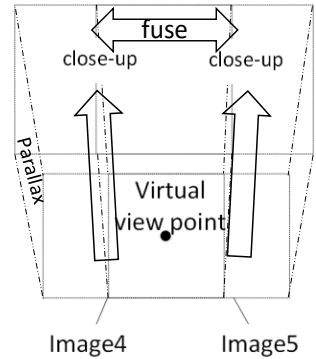

**Figure 3.** A sketch map of close-up viewpoint.

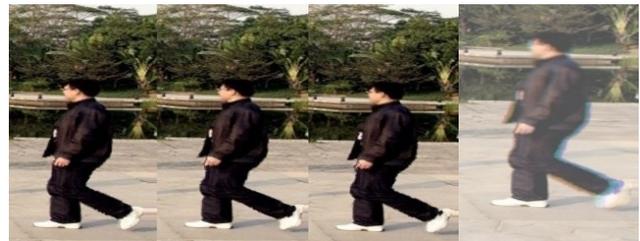

(a) View point 1  (b) View point 2  (c) View point 3
(d) Overlapped result

**Figure 4.** Virtual viewpoint results.

## 4. EXPERIMENT RESULTS

### 4.1. Results of virtual view synthesis

In this paper, bidirectional optical flow is used to project the left and right images simultaneously. Optical flow and weight of the pixels are used to generate a particular virtual viewpoint. This method cope with the margin deformation and holes of image problems thus effectively giving high quality views. Compared with other virtual viewpoint methods, our method can be characterized as simple and easy to implement.

The proposed method has been implemented on the two images taken from one camera at different view angles, and we set the ratio of optical flow as: a = 0.25,0.5,0.75. By applying the interpolation of optical flow, we got three viewpoint results as shown in the Figure 4.

(a), (b) and (c) are from three virtual viewpoints. To see more intuitive changes from the perspective view transfer, we change the RGB of three images as (a): R=255; (b): G=255; (c): B=255 respectively. And then magnifying and overlapping them together in (d), the three different colors at the edge of his left leg and right hand can be clearly seen, which shows the changes of perspectives.

### 4.2. Close-up virtual views

For close-up scenes, we use the optical flow to estimate the depth and then get lens stretching effect along image magnification to construct a pseudo 3D projection. Thus completely deal with the problem that the virtual image can be only confined to the same plane and partially removing the magnification distortion of the image.

In Figure 5. The proposed method is compared with simple magnification, to make comparison more obvious, we took simple magnification result and also adjust the transparency of our close–up result as 50% to get the overlapped image. And from the image it can be seen clearly that only the foreground is changed with little affecting the background.

From comparison of all the experiments, we can see that the proposed method achieves the best magnification quality with more occlusion between foreground and background through the lens stretching process, which makes it more stereoscopic.

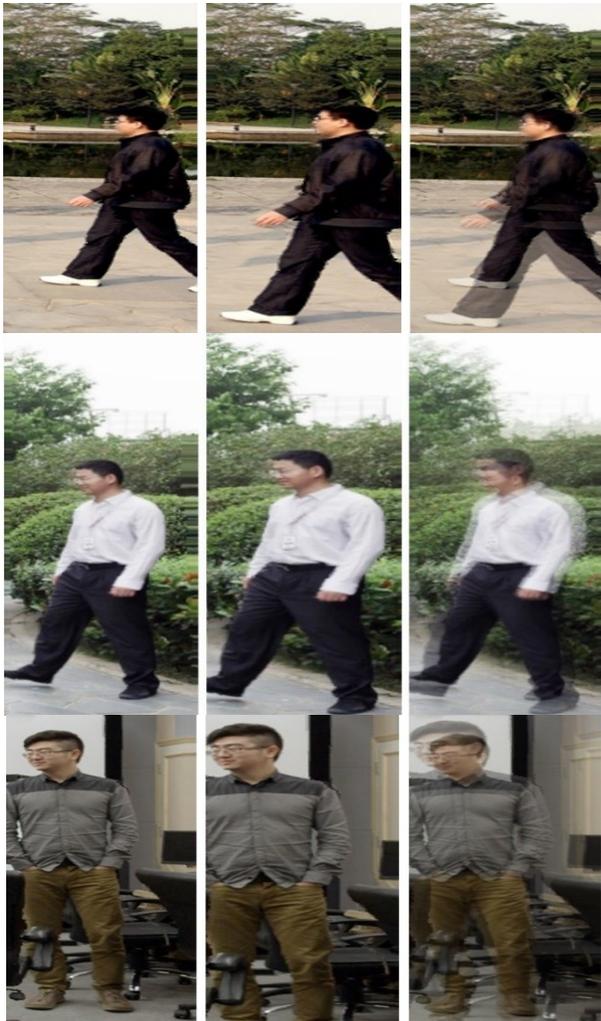

(a) Simple magnification  (b) Close-up result  (c) Overlapped results

**Figure 5.** Our close-up results compared with simple magnification.

### 5. CONCLUSIONS

In this paper, optical flow is used for the close-up image synthesis in image-based rendering. We have successfully used our close-up view synthesis algorithm to establish a pseudo 3D projection magnification based only on optical flow. Our magnification results are obvious in lens stretch that give us a much stronger realistic and stereo feeling than simple magnification. The proposed interpolation of optical flow has been successfully applied to generate a clear scene from any virtual viewpoint, which can remove distortion and visual unreality.

### 6. REFERENCES

[1] H. Shum, S. Kang, "Review of image-based rendering techniques." *Visual Communications and Image Processing*, 2000.

[2] S. Kolhatkar, R. Laganiere, "Real-Time Virtual Viewpoint Generation on the GPU for Scene Navigation." *Canadian Conference on Computer and Robot Vision*, 2010.

[3] Y. Furukawa, et, al. "Reconstructing building interiors from images." *International Conference on Computer Vision*, 2009.

[4] B. Horn, B. Schunck, "Determining Optical Flow." *Artificial Intelligence*, 17(1-3): 185-203, 1993.

[5] T. Liu, "The study and application of video vehicles detection and tracking based on optical flow field ". *Thesis Wuhan University of Science and Technology*, 2011.

[6] Y. Niu, A. Dick, M. Brooks, "Locally Oriented Optical Flow Computation." *IEEE Transactions on Image Processing*, 21(4): 1573-1586, 2012.

[7] Z. Hou, C. Han, "A Background Reconstruction Algorithm based on Pixel Intensity Classification in Remote Video Surveillance System1.", 2001.

[8] B. Lucas, T. Kanade, "An iterative image registration technique with an application to stereo vision." *International Joint Conference on Artificial Intelligence*, 1981.

[9] M. Faisal, J. Barron, "High accuracy optical flow method based on a theory for warping: implementation and qualitative/quantitative evaluation." *International Conference on Image Analysis and Recognition*, 2007.

[10] C. Liu, "Beyond pixels: Exploring new representations and applications for motion analysis". *Massachusetts Institute of Technology*, 2009.

[11] V. J. Tsai, C.T. Chang, "Three-dimensional positioning from Google street view panoramas." *IET Image Processing*, 7(3): 229-239, 2013.

[12] D. Capel, A. Zisserman., "Computer vision applied to super resolution." Signal Processing M*agazine*, *IEEE*, 20(3): 75-86, 2003.

[13] S. P. Kaufman, A. Savikovsky, "3D projection with image recording." *U.S. Patent* No. 6,935,748. 30, 2005.

[14] L. Yang, et, al. "Probabilistic reliability based view synthesis for FTV." *International Conference on Image Processing*, 2010.

[15] W. Xu, J. Mulligan, "Performance evaluation of color correction approaches for automatic multi-view image and video stitching" *Computer Vision and Pattern Recognition (CVPR)*, *IEEE Conference on. IEEE,* 2010.